\documentclass[graybox]{svmult}

\usepackage{mathptmx}       % selects Times Roman as basic font
\usepackage{helvet}         % selects Helvetica as sans-serif font
\usepackage{courier}        % selects Courier as typewriter font
\usepackage{type1cm}        % activate if the above 3 fonts are
\usepackage{makeidx}         % allows index generation
\usepackage{graphicx}        % standard LaTeX graphics tool
\usepackage{multicol}        % used for the two-column index
\usepackage[bottom]{footmisc}% places footnotes at page bottom
\usepackage[hyphens]{url}
\usepackage{hyperref}
\usepackage[hyphenbreaks]{breakurl}
\usepackage{ascmac}
\usepackage{tcolorbox}
\tcbuselibrary{raster,skins,breakable}
\tcbuselibrary{xparse} %preamble
\DeclareTColorBox{brekableitembox}{ o m O{.5} O{} }% 
  {empty, left=2mm, right=2mm, top=-2mm, bottom=0.2mm, attach boxed title to top left={xshift=1.2em},
  boxed title style={empty,left=-2mm,right=-2mm}, colframe=black, coltitle=black, coltext=black, breakable,  
  underlay unbroken={\draw[black,line width=#3pt]
    (title.east) -- (title.east-|frame.east) -- (frame.south east) -- (frame.south west) -- (title.west-|frame.west) -- (title.west); },
  underlay first={\draw[black,line width=#3pt](title.east) -- (title.east-|frame.east) -- (frame.south east) ;
    \draw[black,line width=#3pt] (frame.south west) -- (title.west-|frame.west) -- (title.west); },
  underlay middle={\draw[black,line width=#3pt](frame.north east) -- (frame.south east) ;
    \draw[black,line width=#3pt](frame.south west) -- (frame.north west) ;},
  underlay last={\draw[black,line width=#3pt](frame.north east) -- (frame.south east) -- (frame.south west) -- (frame.north west) ;}, IfValueTF={#1}{title=~~#2~~〈#1〉}{title=~~#2~~},#4}
\usepackage{ulem}

\makeindex             % used for the subject index
\begin{document}

\title*{llm-japanese-dataset v0: Construction of Japanese Chat Dataset for Large Language Models and its Methodology}
\titlerunning{Construction of Japanese Chat Dataset for LLMs and its Methodology}
% Use \titlerunning{Short Title} for an abbreviated version of
% your contribution title if the original one is too long
\author{Masanori HIRANO, Masahiro SUZUKI, and Hiroki SAKAJI}
% Use \authorrunning{Short Title} for an abbreviated version of
% your contribution title if the original one is too long
\institute{Masanori HIRANO, Masahiro SUZUKI, and Hiroki SAKAJI \at The University of Tokyo, 7--3--1 Hongo, Bunkyo, Tokyo
  113--8656 Japan, \email{research@mhirano.jp, b2019msuzuki@socsim.org, sakaji@sys.t.u-tokyo.ac.jp}}
%
% Use the package "url.sty" to avoid
% problems with special characters
% used in your e-mail or web address
%
\maketitle

\abstract*{This study constructed a Japanese chat dataset for tuning large language models (LLMs), which consist of about 8.4 million records.
Recently, LLMs have been developed and gaining popularity.
However, high-performing LLMs are usually mainly for English.
There are two ways to support languages other than English by those LLMs: constructing LLMs from scratch or tuning existing models.
However, in both ways, datasets are necessary parts.
In this study, we focused on supporting Japanese in those LLMs and making a dataset for training or tuning  LLMs in Japanese.
The dataset we constructed consisted of various tasks, such as translation and knowledge tasks.
In our experiment, we tuned an existing LLM using our dataset and evaluated the performance qualitatively.
The results suggest that our dataset is possibly beneficial for LLMs.
However, we also revealed some difficulties in constructing LLMs in languages other than English.}

\abstract{This study constructed a Japanese chat dataset for tuning large language models (LLMs), which consist of about 8.4 million records.
Recently, LLMs have been developed and gaining popularity.
However, high-performing LLMs are usually mainly for English.
There are two ways to support languages other than English by those LLMs: constructing LLMs from scratch or tuning existing models.
However, in both ways, datasets are necessary parts.
In this study, we focused on supporting Japanese in those LLMs and making a dataset for training or tuning  LLMs in Japanese.
The dataset we constructed consisted of various tasks, such as translation and knowledge tasks.
In our experiment, we tuned an existing LLM using our dataset and evaluated the performance qualitatively.
The results suggest that our dataset is possibly beneficial for LLMs.
However, we also revealed some difficulties in constructing LLMs in languages other than English.}

\section{Introduction}
Large language models (LLMs) have recently achieved remarkable progress in performance and generalization.
Specifically, Transformer-based LLMs such as BERT \cite{Devlin2018} and the GPT series \cite{GPT-1,GPT-2,GPT-3} have demonstrated high-performance thanks to their pre-training.
Furthermore, models that have evolved from these, such as ChatGPT \cite{chatgpt} and GPT4 \cite{gpt4}, have gained popularity for their remarkable performance.
Other models such as Bard \cite{bard}, LLaMA \cite{touvron2023llama}, Dolly \cite{dolly}, Bloom \cite{scao2022bloom}, and Vicuna \cite{vicuna} have also emerged.

Some of those models are already provided to consumers as a web service.
Moreover, via API, those models and services are also now available for sub-parts of web services, and many spin-off services are emerging.

However, despite the prosperity of language models, there are still challenges in handling diverse prompts, including prompts written in languages other than English.
For example, Alpaca \cite{alpaca} dataset has been proposed due to the incompleteness of LLaMA's response.
However, the dataset of Alpaca is only available in English, and the incompleteness pointed out by Alpaca has not been filled yet in the other languages.
Moreover, LLaMA has difficultness to respond appropriately to some prompts in languages other than English.

Considering these challenges, it is necessary to enhance models' performances in languages other than English.
However, it is not a good idea to study a specific model in terms of performance improvements in the other language.
Moreover, model development is still ongoing and very competitive, and the situation is changing dramatically recently.
It is also easy to assume that newer models with better performance will emerge in a few months or even 1-2 months.
Therefore, enhancing datasets that support model training may be more useful than focusing on specific models.
This approach may also lower the barrier to adapting new models to languages other than English.

Therefore, this study constructed a new chat dataset in Japanese for LLM training, which contains approximately 8.4 million data points, and demonstrated the performance of the dataset qualitatively.
The dataset and trained models are open-sourced and publicly available. The details are as follows:
\begin{itemize}
    \item Dataset: \url{https://huggingface.co/datasets/izumi-lab/llm-japanese-dataset}
    \item Trained Models (LLaMA 1 epoch): \\ \url{https://huggingface.co/izumi-lab/llama-13b-japanese-lora-v0-1ep}
\end{itemize}
The more details are explained in the following.
Moreover, data expansion and additional model training are planned as future tasks.

% \section{Related Work}

\section{Dataset Construction: izumi-lab/llm-japanese-dataset v0}
In this study, we created a Japanese chat dataset.
The dataset\footnote{In this paper, we mentioned v0, specifically v0.1.0. There is a possibility of future updates.} contains 8,393,726 data points.
In the following, we describe the details of datasets and their creation process.

The dataset is publicly available at the following URLs:
\begin{itemize}
    \item Hugging Face: \url{https://huggingface.co/datasets/izumi-lab/llm-japanese-dataset}
    \item GitHub: \url{https://github.com/masanorihirano/llm-japanese-dataset}
\end{itemize}

The format of the chat data used for model training is shown below.
In the description of the dataset later, we will omit some of the introductory parts and line breaks.

\begin{brekableitembox}{Format used for chat data during model training}
\small
Below is an instruction that describes a task, paired with an input that provides further context. Write a response that appropriately completes the request.
\\\\
\#\#\# Instruction:\\
\{Instruction\}
\\\\
\#\#\# Input:\\
\{Input\}
\\\\
\#\#\# Response:\\
\{Resoponse\}
\end{brekableitembox}
Note that, in the following examples, the underlined sentences are originally written in Japanese.

\subsection{Translation Task}
In many LLM models, while the performance in English is good, the support for languages other than English, such as Japanese, is often immature.
In such cases, if it were possible to link knowledge in English with other languages, such as Japanese, it would be possible to transfer the knowledge learned in English to multilingual settings.
Therefore, we set up translation tasks from Japanese to English and from English to Japanese.

\newcounter{footnotecommon1}
\footnotetext{Originally written in Japanese.}\setcounter{footnotecommon1}{\value{footnote}}
\begin{brekableitembox}{Translation Task Example 1}
\small
\#\#\# Instruction:\\
\uline{Please translate the following Japanese into English.}\footnotemark[\value{footnotecommon1}]\\
\#\#\# Input:\\
\uline{He doesn't see his family in his busy life.}\footnotemark[\value{footnotecommon1}]\\
\#\#\# Response:\\
He doesn't see his family in his busy life.
\end{brekableitembox}

\begin{brekableitembox}{Translation Task Example 2}
\small
\#\#\# Instruction:\\
\uline{Please translate the title of the following English paper into Japanese.}\footnotemark[\value{footnotecommon1}]\\
\#\#\# Input:\\
Odd viscosity in chiral active fluids\\
\#\#\# Response:\\
\uline{Odd viscosity in chiral active fluids}\footnotemark[\value{footnotecommon1}]
\end{brekableitembox}

The translation data was acquired by processing the following datasets:
\begin{itemize}
    \item {\bf Coursera Corpus Mining and Multistage Fine-Tuning for Improving Lectures Translation} \cite{song-etal-2020-coursera}\footnote{Available at \url{https://github.com/shyyhs/CourseraParallelCorpusMining}}: The translation data was utilized for both Japanese-English and English-Japanse directions using the format of Example 1. The license is Apache License 2.0. The dataset size created is 106,332.
    \item {\bf ParaNatCom --- Parallel English-Japanese abstract corpus made from Nature Communications articles} \cite{ParaNatCom}: Among the translation data related to papers, the title and abstract were utilized for both Japanese-English and English-Japanse directions using the format of Example 2. The license is CC BY 4.0. The dataset size created is 1,732.
    \item {\bf Tab-delimited Bilingual Sentence Pairs}\footnote{Available at \url{http://www.manythings.org/anki/}}: The Japanese-English parallel translation data was used for both Japanese-English and English-Japanse directions using the format of Example 1. The license is CC BY. The dataset size created is 202,398.
    \item {\bf Asian Language Treebank (ALT) Project} \cite{riza2016introduction}\footnote{Available at \url{https://www2.nict.go.jp/astrec-att/member/mutiyama/ALT/}}: Only the Japanese-English parallel translation data was used for both Japanese-English and English-Japanse directions using the format of Example 1. The license is CC BY 4.0. The dataset size created is 39,816.
    \item {\bf Tanaka Corpus}\footnote{Available at \url{http://www.edrdg.org/wiki/indExamplephp/Tanaka_Corpus}}: The Japanese-English parallel translation data was used for both Japanese-English and English-Japanse directions using the format of Example 1. The license is CC BY. The dataset size created is 298,026.
    \item {\bf Japanese-English Subtitle Corpus} \cite{pryzant_jesc_2018}\footnote{Available at \url{https://nlp.stanford.edu/projects/jesc/index_ja.html}}: The Japanese-English parallel translation data was used for both Japanese-English and English-Japanse directions using the format of Example 1. The license is CC BY-SA 4.0. The dataset size created is 5,396,204.
    \item {\bf Japanese WordNet} \cite{isahara2008development}\footnote{Available at \url{https://bond-lab.github.io/wnja/}}: The description and usage examples of words were used as a translation dataset, using the English parallel dataset in the original Wordnet \cite{fellbaum2010wordnet}, for both Japanese-English and English-Japanse directions using the format of Example 1. The license is a BSD-like license unique to it. The dataset size created is 367,936.
    \item {\bf Easy Japanese Corpus}: The Japanese-English parallel data included in ``SNOW T15: Japanese Simplified Corpus with Core'' Vocabulary \cite{maruyama2018simplified} and ``SNOW T23: Crowdsourced Corpus of Sentence Simplification with Core Vocabulary'' \cite{katsuta2018crowdsourced} was used for both Japanese-English and English-Japanse directions using the format of Example 1. The license is CC BY 4.0. The dataset size created is 168,600.
\end{itemize}

Combined, the total number of data is 6,581,044.
Although this occupies approximately 80\% of the dataset size, we decided to adopt all of them due to the motivation mentioned earlier.

\subsection{Knowledge-based Tasks}
Next, to appropriately learn knowledge in Japanese, knowledge-based tasks were incorporated into the training dataset.
In the following, we explain the datasets used for creating our dataset and the created dataset examples.

\begin{itemize}
    \item {\bf Wikipedia}\footnote{Available at \url{https://dumps.wikimedia.org/}. Used jawiki version 20230401}: By using the titles and first paragraphs of Wikipedia, the task of explaining words was created. The license is CC BY-SA 3.0. The dataset size created is 1,347,381.
    \item {\bf Japanese WordNet} \cite{isahara2008development}\footnote{Available at \url{https://bond-lab.github.io/wnja/}}: By using the word and definition parts, a chat was created that requests word explanations. The license is a BSD-like license unique to it. The dataset size created is 173,085.
    \item {\bf AIO (AI King) Official Distribution Dataset Version 2.0}\footnote{Available at \url{https://sites.google.com/view/project-aio/dataset}}: Since it is a quiz format dataset, it was incorporated directly as a chat dataset. The license is CC BY-SA 4.0. The dataset size created is 23,335.
    \item {\bf Japanese Movie Recommendation Dialogue (JMRD)} \cite{kodama2022construction} \footnote{Available at \url{https://github.com/ku-nlp/JMRD}}: Only the metadata related to movies was extracted, and a chat dataset was created that shows movie descriptions and summaries. The license is CC BY-SA 4.0. The dataset size created is 260.
    \item {\bf JCommonsenseQA} \cite{kurihara-etal-2022-jglue}\footnote{Available at \url{https://github.com/yahoojapan/JGLUE}}: A Q\&A dataset that questions common sense, which is part of the Japanese General Language Understanding Evaluation (JGLUE), was incorporated as its original form. The license is CC BY-SA 4.0. The dataset size created is 10,058.
    \item {\bf Aozorra Paperback}\footnote{\url{https://www.aozora.gr.jp/}}: Only works of Japanese literature whose copyright has expired were selected from Aozora Paperback\footnote{Used \url{https://github.com/levelevel/AozoraTxt} for this purpose.}, and questions were created that ask for the beginning of the sentence and the author and title from the beginning of the sentence. The dataset size created is 33,758.
\end{itemize}
Combining them, the total number of datasets is 1,587,877.

\begin{brekableitembox}{Knowledge-based Tasks: Wikipedia}
\small
\#\#\# Instruction:\\
\uline{Please explain the input word.}\footnotemark[\value{footnotecommon1}]\\
\#\#\# Input:\\
\uline{The University of Tokyo}\footnotemark[\value{footnotecommon1}]\\
\#\#\# Response:\\
\uline{The University of Tokyo (Tokyo Daigaku) is a Japanese national university headquartered in Bunkyo-ku, Tokyo. Its abbreviated name is Todai.}\footnotemark[\value{footnotecommon1}]
\end{brekableitembox}

\begin{brekableitembox}{Knowledge-based Tasks: Japanese WordNet}
\small
\#\#\# Instruction:\\
\uline{What does "possible" mean?}\footnotemark[\value{footnotecommon1}]\\
\#\#\# Response:\\
\uline{Have the methods, skills or know-how or rights necessary to do something}\footnotemark[\value{footnotecommon1}]
\end{brekableitembox}

\begin{brekableitembox}{Knowledge-based Tasks: AIO}
\small
\#\#\# Instruction:\\
\uline{What bird is used to describe the personalities of three warlords, Nobunaga Oda, Hideyoshi Toyotomi, and Ieyasu Tokugawa?}\footnotemark[\value{footnotecommon1}]\\
\#\#\# Response:\\
\uline{Cuculus poliocephalus}\footnotemark[\value{footnotecommon1}]
\end{brekableitembox}

\begin{brekableitembox}{Knowledge-based Tasks: JMRD (partially omitted)}
\small
\#\#\# Instruction:\\
\uline{Tell us about My Neighbor Totoro.}\footnotemark[\value{footnotecommon1}]\\
\#\#\# Response:\\
\uline{My Neighbor Totoro is a movie produced in 1988 and directed by Hayao Miyazaki. Noriko Hidaka and Chinatsu Sakamoto appear in the movie. The summary is as follows:}\\
\uline{[Summary]}\\
\uline{An elementary school student Satsuki and her younger sister Mei move with their father to a farming village in early summer for their mother's medical treatment. ...}\footnotemark[\value{footnotecommon1}]
\end{brekableitembox}

\begin{brekableitembox}{Knowledge-based Tasks: JCommonsenseQA}
\small
\#\#\# Instruction:\\
\uline{Which of the following are primarily for children and contain illustrated stories?}\footnotemark[\value{footnotecommon1}]\\
\#\#\# Input:\\
\uline{World, photo book, illustrated book, research paper, picture book}\footnotemark[\value{footnotecommon1}]\\
\#\#\# Response:\\
\uline{Illustrated book}\footnotemark[\value{footnotecommon1}]
\end{brekableitembox}

\begin{brekableitembox}{Knowledge-based Tasks: Aozorra Paperback (1)}
\small
\#\#\# Instruction:\\
\uline{What is the beginning of Ivan Turgenev's work ``First Love''?}\footnotemark[\value{footnotecommon1}]\\
\#\#\# Response:\\
\uline{It begins with ``The guests had long since departed. The clock struck half-past twelve. There remained in the room only the host, Sergyéi Nikoláevitch, and Vladímir Petróvitch. The host rang and ordered the remains of the supper to be removed.''}\footnotemark[\value{footnotecommon1}]
\end{brekableitembox}

\begin{brekableitembox}{Knowledge-based Tasks: Aozorra Paperback (2)}
\small
\#\#\# Instruction:\\
\uline{Please tell us the author and title of the literary work that begins, ``Once there were two young gentlemen climbing up deep into the mountains. They dressed like British soldiers and walking on the dry leaves carrying shining guns on their shoulders.''}\footnotemark[\value{footnotecommon1}]\\
\#\#\# Response:\\
\uline{``The Restaurant of Many Orders'' written by Kenji Miyazawa.}\footnotemark[\value{footnotecommon1}]
\end{brekableitembox}

\subsection{Other Tasks}
\subsubsection{Creating a Paper Title and Imagining an Abstract}
We utilized the aforementioned ParaNatCom \cite{ParaNatCom} to create tasks related to our research paper.
The license for the dataset is CC BY 4.0, and the size of the created dataset is 1,732.
\begin{brekableitembox}{Task Example 1 (partially omitted)}
\small
\#\#\# Instruction:\\
\uline{Please make a title from the abstract of the paper.}\footnotemark[\value{footnotecommon1}]\\
\#\#\# Input:\\
\uline{Superthin nanostructures, particularly with atomic-level thicknesses, typically display unique optical properties because of their exceptional light-matter interactions. Here, we report a facile strategy for the synthesis of sulfur-doped molybdenum oxide nanorings ...}\footnotemark[\value{footnotecommon1}]\\
\#\#\# Response:\\
\uline{Atomic-level molybdenum oxide nanorings with full-spectrum absorption and photoresponsive properties}\footnotemark[\value{footnotecommon1}]
\end{brekableitembox}

\begin{brekableitembox}{Task Example 2 (partially omitted)}
\small
\#\#\# Instruction:\\
\uline{Imagine and write an abstract from the paper's title.}\footnotemark[\value{footnotecommon1}]\\
\#\#\# Input:\\
\uline{Surface plasmon induced direct detection of long wavelength photons}\footnotemark[\value{footnotecommon1}]\\
\#\#\# Response:\\
\uline{Millimeter and terahertz wave photodetectors have long been of great interest due to a wide range of applications, but they still face challenges in detection performance. Here, we propose a new strategy for the direct detection of millimeter and terahertz wave photons based on localized surface-plasmon-polariton (SPP)-induced non-equilibrium ...}\footnotemark[\value{footnotecommon1}]
\end{brekableitembox}

\subsubsection{News Title Generation}
We retrieved articles from the Japanese version of Wikinews\footnote{Crawling from \url{https://ja.wikinews.org/wiki/}}, and set the task of generating a title from the article body.
The license is CC BY 2.5.
The size of the created dataset is 4,265.

\begin{brekableitembox}{Task Example (partially omitted)}
\small
\#\#\# Instruction:\\
\uline{Please write a title for the following news article.}\footnotemark[\value{footnotecommon1}]\\
\#\#\# Input:\\
\uline{On January 11, Iwanami Shoten released the 6th edition of Kojien, the first major revision of the Kojien dictionary in 10 years. The sixth edition will be the first revision in 10 years ...}\footnotemark[\value{footnotecommon1}]\\
\#\#\# Response:\\
\uline{Kojien is revised for the first time in 10 years - Sixth edition released.}\footnotemark[\value{footnotecommon1}]
\end{brekableitembox}

\subsubsection{Answer Extraction Task}
In addition to knowledge and translation tasks, we also adopted a task of extracting answers from the given context.
As the source data, we used JSQuAD \cite{kurihara-etal-2022-jglue}\footnote{Available at \url{https://github.com/yahoojapan/JGLUE}}, which is part of JGLUE tasks.
The license is CC BY-SA 4.0, and the dataset size created is 67,491.

\begin{brekableitembox}{Task Example (partially omitted)}
\small
\#\#\# Instruction:\\
\uline{When did the smallpox pandemic occur?}\footnotemark[\value{footnotecommon1}]\\
\#\#\# Input:\\
\uline{Emperor Shomu: In 737, a smallpox epidemic broke out, resulting in the disastrous deaths of most of the high government officials, including the four Fujiwara brothers. Furthermore, in 740, the Fujiwara no Hirotsugu ...}\footnotemark[\value{footnotecommon1}]\\
\#\#\# Response:\\
\uline{In 737}\footnotemark[\value{footnotecommon1}]
\end{brekableitembox}

\subsubsection{Easy Japanese Task}
We incorporated the task of rewriting normal Japanese into simpler Japanese.
We processed SNOW T15: Japanese Simplified Corpus with Core Vocabulary \cite{maruyama2018simplified} and SNOW T23: Crowdsourced Corpus of Sentence Simplification with Core Vocabulary \cite{katsuta2018crowdsourced} as the original data.
The license is CC BY 4.0.
The size of the created dataset is 84,300.

\begin{brekableitembox}{Task Example}
\small
\#\#\# Instruction:\\
\uline{Please rephrase the following Japanese into easy Japanese.}\footnotemark[\value{footnotecommon1}]\\
\#\#\# Input:\\
\uline{Bill has no sense of adventure at all.}\footnotemark[\value{footnotecommon1}]\\
\#\#\# Response:\\
\uline{Bill has no desire to do anything dangerous.}\footnotemark[\value{footnotecommon1}]
\end{brekableitembox}

\subsubsection{Japanese Translation of Existing Chat Dataset}
In addition, we incorporated Japanese-translated versions of existing publicly available chat datasets.
The following datasets were included:
\begin{itemize}
    \item Japanese-Alpaca-LoRA\footnote{Available at \url{https://github.com/masa3141/japanese-alpaca-lora}}: A translation of the Alpaca\cite{alpaca} dataset into Japanese. The license is Apache License 2.0. The dataset size is 52,002.
    \item databricks-dolly-15k-ja\footnote{Available at \url{https://github.com/kunishou/databricks-dolly-15k-ja}}: A Japanese-translated version of the dataset used for training Dolly\cite{dolly}. The license is CC BY-SA 3.0. The dataset size is 15,015.
\end{itemize}

\section{Model Construction and Evaluation Using Our Dataset}
This study used LoRA \cite{hu2022lora} as a method to fine-tune LLMs without significant performance degradations.
It is because building LLMs from scratch requires a massive amount of computational resources.
Furthermore, LLMs with a large number of parameters require GPU resources not only for pre-training but also for fine-tuning.
On the other hand, LoRA updates only small parts of LLM parameters.
Therefore, LoRA is a feasible option for us to evaluate the benefits of our dataset.

The main parameters used in the experiment are shown below.
\begin{itemize}
  \item Base model: LLaMA 13B \cite{touvron2023llama}
  \item Learning rate: 3e-4
  \item Input length: 256 tokens
  \item Batch size: 130
  \item Epochs: 1
  \item $r$ of LoRA: 4
  \item $\alpha$ of LoRA: 16
  \item Dropout rate of LoRA: 0.05
  \item Tuning parameters in LoRA: q\_proj, v\_proj
\end{itemize}
We used PEFT\cite{PEFT} and DeepSpeed ZeRO 2\cite{rajbhandari2020zero} for the implementation.

This tuned model is publicly available at \url{https://huggingface.co/izumi-lab/llama-13b-japanese-lora-v0-1ep}.

In order to increase the reproducibility of the evaluation experiment, the temperature parameter for prompt generation was set to 0.0.
Below are some qualitative comparisons we conducted to assess performance.

\begin{brekableitembox}{Response Example (1)}
\small
\#\#\# Input:\\
\uline{Who are you?}\footnotemark[\value{footnotecommon1}]\\
\#\#\# Output(LLaMA+LoRA):\\
\uline{I am human.}\footnotemark[\value{footnotecommon1}]\\
\#\#\# Output(LLaMA):\\
\uline{This application uses Google items.}\footnotemark[\value{footnotecommon1}]
\end{brekableitembox}

\begin{brekableitembox}{Response Example (2)}
\small
\#\#\# Input:\\
\uline{How can I get to Osaka from Tokyo?}\footnotemark[\value{footnotecommon1}]\\
\#\#\# Output(LLaMA+LoRA):\\
\uline{To get to Osaka from Tokyo, take the train from Tokyo Station to Osaka Station.}\footnotemark[\value{footnotecommon1}]\\
\#\#\# Output(LLaMA):\\
\uline{How can I get to Osaka from Tokyo.}\footnotemark[\value{footnotecommon1}]
\end{brekableitembox}

\begin{brekableitembox}{Response Example (3)}
\small
\#\#\# Input:\\
\uline{Where is the University of Tokyo located?}\footnotemark[\value{footnotecommon1}]\\
\#\#\# Output(LLaMA+LoRA):\\
\uline{The University of Tokyo is located in Tokyo, Japan.}\footnotemark[\value{footnotecommon1}]\\
\#\#\# Output(LLaMA):\\
\uline{The University of Tokyo is located at 1-1-1 Bunkyo, Bunkyo-ku, Tokyo.}\footnotemark[\value{footnotecommon1}]\\
(Authors' note: The correct address is 7-3-1 Hongo, Bunkyo-ku, Tokyo.)
\end{brekableitembox}

\begin{brekableitembox}{Response Example (4)}
\small
\#\#\# Input:\\
\uline{A phone rang. What should I do?}\footnotemark[\value{footnotecommon1}]\\
\#\#\# Output(LLaMA+LoRA):\\
\uline{To hang up the phone.}\footnotemark[\value{footnotecommon1}]\\
\#\#\# Output(LLaMA):\\
\uline{The phone rings. When the call is received, the person receiving the call should receive the call.}\footnotemark[\value{footnotecommon1}]
\end{brekableitembox}

\begin{brekableitembox}{Response Example (5)}
\small
\#\#\# Input:\\
\uline{What are the three major festivals in Kyoto?}\footnotemark[\value{footnotecommon1}]\\
\#\#\# Output(LLaMA+LoRA):\\
\uline{The three major festivals in Kyoto are the spring festival, the summer festival, and the autumn festival.}\footnotemark[\value{footnotecommon1}]\\
(Authors' note: Correct answer is Aoi, Gion, Jidai festivals.)\\
\#\#\# Output(LLaMA):\\
\uline{What are the three major festivals in Kyoto?}\\
\uline{What are the three major festivals in Kyoto?}\\
\uline{What are the three major festivals in Kyoto?}\footnotemark[\value{footnotecommon1}]
\end{brekableitembox}

\section{Discussion}
In this study, we constructed a dataset, izumi-lab/llm-japanese-dataset v0.
The dataset consists of approximately 8.4 million data points, mainly focusing on Japanese-English translation tasks.
The purpose of this dataset was to improve the performance of Japanese language processing capability and connect English language processing capabilities with Japanese input and output.
Therefore, we tested additional LoRA tuning to the LLM using our datasets to evaluate its benefits.

Reviewing the actual prompt responses after LoRA tuning, there appears to be a slight improvement in performance.
However, since we only trained LLaMA with one epoch of LoRA tuning, the difference may be small.
Therefore, increasing the number of epochs could be a future work. Nevertheless, the results suggest that our new dataset is effective.
For example, the responses of our model (LLaMA + LoRA) in examples (4) and (5) were incorrect, but LoRA tuning seems to achieve generating more appropriate Japanese outputs.
Additionally, although the responses of our model (LLaMA + LoRA)  in examples (2) and (3) were still somewhat unfluent, they have achieved generated correct responses in terms of facts.

However, this qualitative comparison is not fully appropriate, and establishing a comparison method is also a possible future work.
To evaluate more accurately, it is necessary to construct evaluation tasks and datasets specialized for Japanese, and there may be a need for a Japanese version of the Language Model Evaluation Harness \cite{eval-harness} evaluation scheme.

\section{Conclusion}
In this study, we constructed a Japanese chat dataset for training LLMs on chat responses.
The dataset can be used for tuning chat tasks to LLMs created mainly in English, and we qualitatively confirmed the performance of improvement in the LLM tuned by our dataset and LoRA tuning.
Our dataset is now publicly available.
Through the construction of the model to check the performance of the dataset, we have also clarified some issues in the construction of LLMs for languages other than English.
Such a trial to adapt LLMs built mainly for English into other languages is now essential to make LLMs available fluently in wider languages.

\begin{acknowledgement}
This work was supported in part by JSPS KAKENHI Grant Number JP21K12010 and JST PRESTO Grant Number JPMJPR2267.
\end{acknowledgement}

\bibliographystyle{spmpsci}
\bibliography{cite}
\end{document}